\documentclass[10pt,a4paper,logo]{paper}

\usepackage[all]{hypcap}

\usepackage[authoryear, round]{natbib}

\usepackage{nicefrac}       
\usepackage{multirow}
\usepackage{cleveref}
\usepackage{dsfont}
\usepackage{makecell}
\usepackage{subcaption}
\usepackage{bbm}
\usepackage{wrapfig}
\usepackage{mathtools,xparse}
\usepackage{array}
\usepackage{arydshln}
\usepackage{scalerel}
\usepackage{tablefootnote}
\usepackage{tocloft}
\usepackage{adjustbox}
\usepackage{setspace}
\usepackage[most,skins,theorems]{tcolorbox}
\usepackage{xspace}
\usepackage[bottom]{footmisc}
\usepackage{indentfirst}
\titlespacing{\paragraph}{\parindent}{0pt}{0.5em}

\newboolean{showsection}
\setboolean{showsection}{true}
\makeatletter
\@namedef{ver@everyshi.sty}{}
\makeatother

\definecolor{custom_green}{rgb}{0.0, 0.5, 0.0}
\definecolor{custom_red}{rgb}{1.0, 0.01, 0.24}
\definecolor{custom_blue}{HTML}{C9DAF7}
\definecolor{custom_purple}{HTML}{D9D1E9}
\definecolor{title_blue}{HTML}{204899}
\definecolor{cite_blue}{HTML}{044dc1}
\definecolor{cite_purple}{HTML}{7406a7}

\hypersetup{
    colorlinks = true,
    citecolor = {cite_blue},
    linkcolor = {cite_purple},
    urlcolor = {cite_purple},
}

\newcommand{\eg}{\emph{e.g.},\ }
\newcommand{\ie}{\emph{i.e.},\ }

\definecolor{blanchedalmond}{rgb}{1.0, 0.92, 0.8}
\definecolor{carmine}{rgb}{0.59, 0.0, 0.09}
\definecolor{lightblue}{rgb}{0.22,0.45,0.70}%

\newtheorem{theorem}{Theorem}[section]

\newtheorem{remark}[theorem]{Remark}

\renewcommand{\mathbf}{\boldsymbol}

\makeatletter
\def\Ddots{\mathinner{\mkern1mu\raise\p@
\vbox{\kern7\p@\hbox{.}}\mkern2mu
\raise4\p@\hbox{.}\mkern2mu\raise7\p@\hbox{.}\mkern1mu}}
\makeatother

\numberwithin{equation}{section}

\definecolor{amaranth}{rgb}{0.9, 0.17, 0.31}
\definecolor{antiquebrass}{rgb}{0.8, 0.58, 0.46}
\definecolor{antiquefuchsia}{rgb}{0.57, 0.36, 0.51}
\definecolor{chromeyellow}{rgb}{0.31, 0.47, 0.26}

\newcommand{\1}{\mathds 1}

\usepackage{amsmath,amsfonts,bm}

\def\eqref#1{equation~\ref{#1}}

\def\1{\bm{1}}

\DeclareMathAlphabet{\mathsfit}{\encodingdefault}{\sfdefault}{m}{sl}
\SetMathAlphabet{\mathsfit}{bold}{\encodingdefault}{\sfdefault}{bx}{n}

\definecolor{cornellred}{HTML}{B31B1B}
\definecolor{jzeroblue}{HTML}{3B75C3}
\definecolor{jzeromaroon}{HTML}{800000}
\newcommand{\ours}{\textsc{J-Zero}\xspace}

\makeatletter
\def\mathcolor#1#{\@mathcolor{#1}}
\def\@mathcolor#1#2#3{%
  \protect\leavevmode
  \begingroup
    \color#1{#2}#3%
  \endgroup
}
\makeatother

\Crefformat{equation}{#2Eq.\;(#1)#3}
\Crefformat{figure}{#2Figure #1#3}
\Crefformat{assumption}{#2Assumption #1#3}
\Crefname{assumption}{Assumption}{Assumptions}

\usepackage{crossreftools}
\makeatletter
\renewcommand\footnoterule{%
  \kern 15\p@
  \hrule \@width 2in \kern 2.6\p@
  \vspace{4pt}
}
\makeatother

\title{J-Zero: Unified Challenger--Solver--Judge \\ Self-Evolution from Zero Data}

\reportnumber{}

\author[1\footnotesize{\textbf{*}}]{Gyouk Chu}
\author[1\footnotesize{\textbf{*}}]{Myeongho Jeon}
\author[2]{Teresa Yeo}
\author[1]{Eunho Yang}

\affil[1]{KAIST}
\affil[2]{Google DeepMind}

\paperlinks{%
\href{https://gyoukchu.github.io/projects/j_zero/}{ 
  \raisebox{-0.2ex}{\includegraphics[height=0.9em]{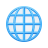}}\,
  \textcolor{RoyalBlue}{Project Page}
}
\hspace{0.8em}
\href{https://github.com/GyoukChu/J-Zero}{ 
  \raisebox{-0.2ex}{\includegraphics[height=0.9em]{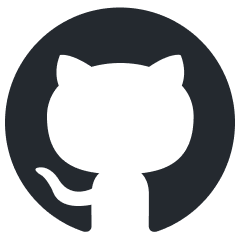}}\,
  \textcolor{RoyalBlue}{GitHub}
}
\hspace{0.8em}
\href{https://huggingface.co/collections/ChuGyouk/j-zero}{ 
  \raisebox{-0.2ex}{\includegraphics[height=0.9em]{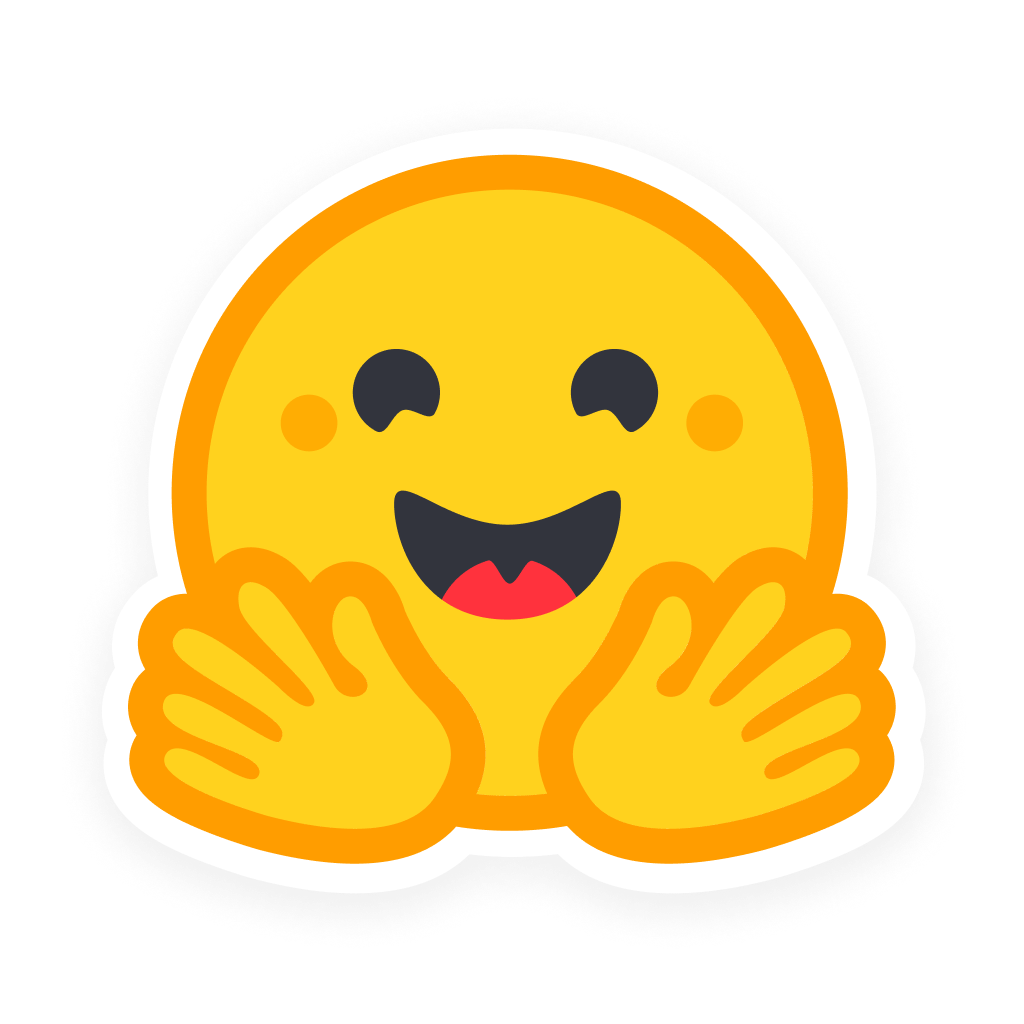}}\,
  \textcolor{RoyalBlue}{Hugging Face}
}
}

\correspondingauthor{\textsuperscript{\textbf{*}}Equal contribution. Correspondence to: \email{eunhoy@kaist.ac.kr}.}

\begin{abstract}
Self-evolving language models have recently emerged as a promising path toward superintelligence, with the advantage of reducing the cost of human supervision.
While considerable progress has been made in verifiable domains, self-evolution in unverifiable domains remains less explored.
We propose \textsc{\textbf{J}}udge co-adaptation from \textsc{\textbf{Zero}} data (\textbf{\ours}), a unified Challenger--Solver--Judge self-evolution framework that supports self-improvement across both domains.
The Challenger and Solver co-evolve through an adversarial interaction: the Challenger generates increasingly difficult tasks, while the Solver learns to produce higher-quality responses to them.
In parallel, the Judge co-adapts using preference pairs whose ordering is known in advance from how each response was produced, \ie the Solver's answer over the Challenger's, and the Solver's decomposed-and-recombined answer over its one-shot answer, rather than from the Judge's own scores.
\ours outperforms the baselines by an average of 4.2 points on verifiable and 8.0 points on unverifiable domains, and continues to improve through at least ten iterations, whereas the baselines degrade after two.
Further analysis identifies Judge co-adaptation as the key driver of this sustained improvement.
\end{abstract}

\begin{document}

\maketitle

\section{Introduction}
\label{sec:intro}


Self-evolving large language models (LLMs) have emerged as a promising approach to overcoming the limitations of human-curated supervision. Relying on human annotators to design tasks and provide labels is costly and constitutes a fundamental bottleneck to developing AI systems that may eventually surpass human intelligence~\citep{tao2024survey, jeon2025weaktostrong}.

Recent work has explored self-evolving models that operate \textit{without any external data}, generating training tasks entirely through closed-loop self-play~\citep{huang2025rzero}. Starting from a single model, these methods instantiate Challenger and Solver roles that co-evolve: the Challenger generates increasingly difficult tasks, while the Solver learns to solve them. Although self-evolving algorithms have been well established in verifiable domains~\citep{acikgoz2026toolr0, yue2026drzero, li2026r}, their application to unverifiable domains remains underexplored.

Self-evolution is relatively straightforward in verifiable domains, where objective ground-truth answers provide direct evaluation and learning signals.
In contrast, unverifiable domains admit no single correct answer, and quality is defined by human preference rather than by a checkable condition.
In this setting, the learning signal comes from a Judge that scores responses in place of a verifier~\citep{kuba2025lsp}. This substitution introduces a ceiling.
A frozen Judge can only push the Solver toward preferences it has already internalized, so once the Solver saturates the distinctions the Judge is able to make, further training yields no signal.
Thus, the extent of self-improvement is bounded by the Judge's own evaluation capability~\citep{huang2026gzero}.
In this regard, we propose a novel framework, \underline{\textsc{\textbf{J}}}udge co-adaptation from \underline{\textsc{\textbf{Zero}}} data (\textbf{\ours}), in which the Judge model \textit{co-adapts} alongside the Challenger and Solver (Figure~\ref{fig:overview}), lifting this ceiling as training proceeds, thereby enabling sustained self-evolution in both verifiable and unverifiable domains. 
The Challenger and Solver co-evolve through a minimax game using group relative policy optimization~\citep[GRPO;][]{shao2024deepseekmath}: the Challenger is trained to minimize the reward assigned by the Judge model by generating increasingly difficult tasks, while the Solver is trained to maximize the reward by producing high-quality responses.
Training the Judge inside this same loop appears circular, \ie if every signal originates from a single model, it is unclear what new preference information could enter the system. Thus, we derive preferences from \textit{structural asymmetries} in the loop: configurations in which one response is better than another by construction.
Specifically, we construct two types of preference pairs for Bradley--Terry~\citep[BT;][]{bradley1952rank}-based Judge training: 
(1) \textit{Role-asymmetry pairs:} the Solver’s response is preferred over the Challenger’s response because the Solver is explicitly optimized to answer the generated task well, whereas the Challenger is optimized to make the task difficult rather than to produce a high-quality answer; and (2) \textit{Subtask-amplification pairs:} the Solver’s divide-and-conquer response is preferred over its one-shot response because solving each subtask accurately is easier than solving the original task as a whole, and aggregating the subtask solutions enables the Solver to produce a more comprehensive and higher-quality response~\citep{christiano2018supervising, wu2021recursively, zhouleast}.

\ours achieves substantial performance improvements across both verifiable and unverifiable domains.
\ours improves average accuracy by \textbf{4.2 points} over the alternative baseline across eleven verifiable benchmarks (Table~\ref{tab:main_results_verifiable}) and improves performance by \textbf{8.0 points} across three benchmarks covering unverifiable tasks (Table~\ref{tab:main_results_unverifiable}).
Beyond these performance gains, our analysis identifies Judge co-evolution as a key component: compared with a frozen Judge, it yields stronger overall performance (Table~\ref{tab:ablation_study}) and sustains improvement across iterative rounds, while the baselines plateau early (Figure~\ref{fig:analysis_lifelong}, \ref{fig:judge_rmbench}), suggesting broader implications for continual and lifelong learning.

\section{Related Work}
\label{sec:related}



\textbf{Self-evolution with external tasks and supervision.} Early self-evolving methods primarily focused on iteratively improving the Solver using tasks paired with ground-truth labels in verifiable domains, aiming to make the most effective use of the available data by adapting the training process to the model’s current capabilities~\citep{zelikman2022star, yuan2023scaling, singh2023beyond, zhang2024restmcts, pang2024iterative}.
Such methods remain bounded by the availability and scope of human-provided labels, offering no direct path to improvement beyond the existing supervision.

\textbf{Self-evolution with external seed resources.} 
A subsequent line of research reduced reliance on ground-truth labels but still depended on external resources. Self-play fine-tuning compares the model’s own responses with reference responses drawn from a supervised fine-tuning (SFT) corpus~\citep{chen2024spin}.
Self-rewarding methods use the model as both the policy and the judge on prompts generated from an external seed dataset~\citep{yuan2024selfrewarding, prasad2024scpo, wang2025cream, wu2025metarewarding, zhou2025scir, wang2025temporal, zhang2025process}, while more recent approaches mine new tasks from raw external corpora~\citep{liu2025spice, huang2026pop, fan2026darc}.
In each case, the scope of self-evolution remains anchored to the initial resource, limiting the amount of genuinely new learnable information and potentially reinforcing the model's existing biases~\citep{liu2026selfevolve}.



\textbf{Data-free self-evolution.}
Zero-data self-play frameworks remove this dependence entirely and differ primarily in how they obtain rewards.
Absolute Zero~\citep{zhao2026absolute} verifies self-proposed coding tasks using an executor, and related work extends the same execution-based feedback to software engineering~\citep{wei2025ssr}.
R-Zero~\citep{huang2025rzero} replaces an external oracle with majority voting over sampled responses, and several successors adopt this strategy for tool use and other settings~\citep{acikgoz2026toolr0, yue2026drzero, li2026r}.
These reward signals are inexpensive to compute and relatively difficult to exploit, but they are largely restricted to verifiable domains. 
\citet{kuba2025lsp} extends data-free self-evolution to unverifiable domains, but its reliance on a static Judge may impose an upper bound on further improvement~\citep{huang2026gzero}.

This leaves one setting unaddressed: data-free self-evolution in which the evaluation signal is itself learned and continually improves across both verifiable and unverifiable domains. To address this, we develop a unified framework that enables such adaptive evaluation.
Concurrent with our work, G-Zero~\citep{huang2026gzero} also extends data-free self-evolution beyond verifiable domains by addressing the limitation that a fixed Judge can cap further improvement. 
Instead of using a fixed Judge, it uses Challenger-generated hints to construct preference pairs between Solver responses and trains the Solver via direct preference optimization (DPO). In contrast, our framework allows the Judge to co-evolve directly with the Challenger and Solver, enabling the evaluation signal itself to improve over successive rounds and thereby leading to more stable and sustained self-improvement.

\section{Methodology}
\label{sec:method}

\begin{figure}[t]
\centering
\includegraphics[width=\linewidth]{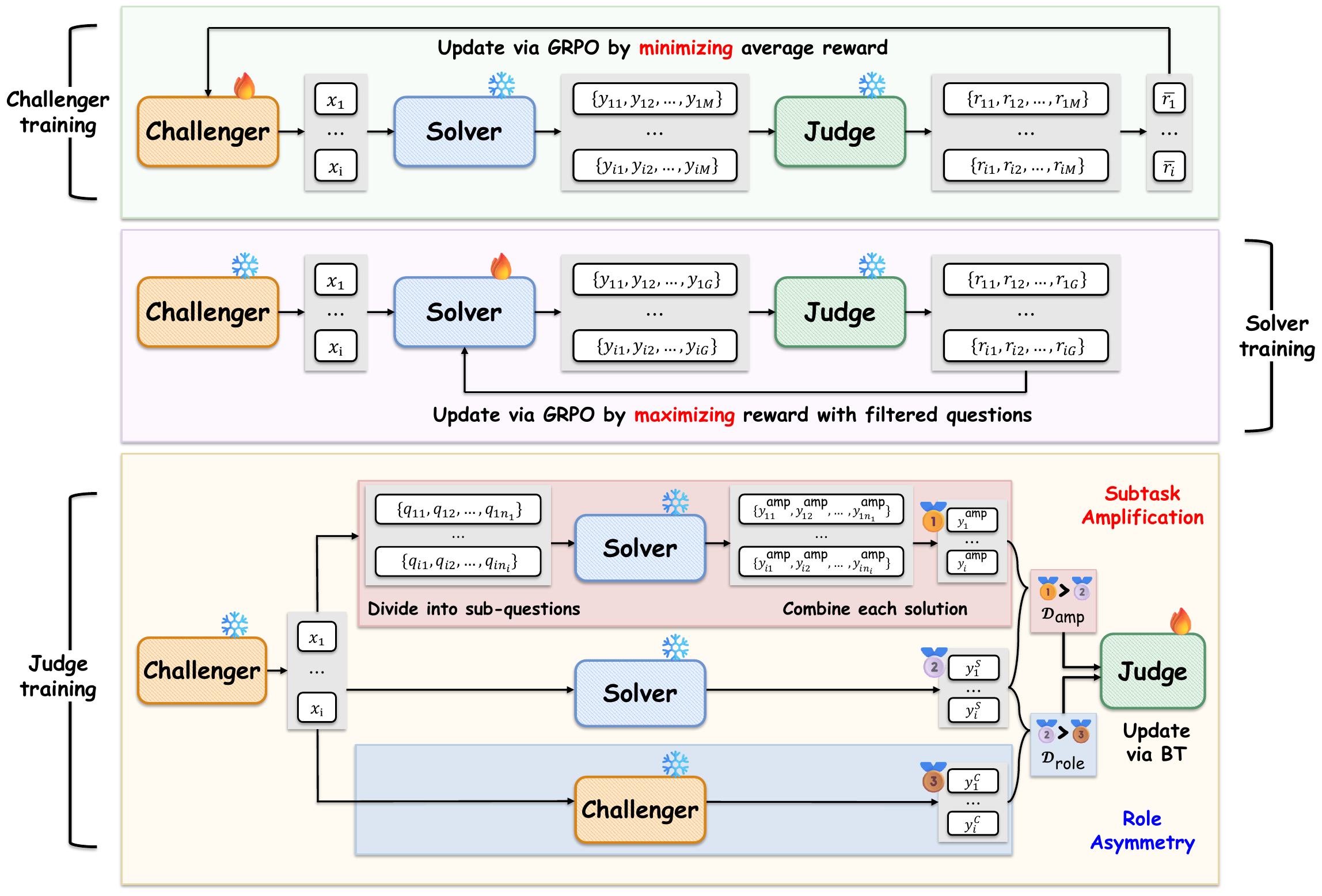}
\caption{
    An overview of \ours, in which the Challenger, Solver, and Judge are jointly updated
    through iterative self-play.
    \textbf{Top and Middle:} In the Challenger and Solver training phases, the two policies are trained adversarially under the frozen Judge. The Challenger generates tasks on which the Solver scores poorly, and the Solver learns to recover high scores on those tasks.
    \textbf{Bottom:} In the Judge training phase, the Judge is updated on two types of in-loop preference pairs, role asymmetry ($\mathcal{D}_{\mathrm{role}}$) and subtask amplification ($\mathcal{D}_{\mathrm{amp}}$), so that its evaluation ability rises in step with the two policies it supervises.
}
\label{fig:overview}
\end{figure}

The use of reward models is a de facto standard for LLM post-training in unverifiable domains~\citep{ouyang2022training}, and recent work has also demonstrated their effectiveness in verifiable domains~\citep{su2025crossing}. However, as noted by \citet{huang2026gzero}, relying on a fixed reward model may impose an upper bound on the overall improvement achievable through self-evolution.

To mitigate this, we propose \ours, a self-evolving framework that co-adapts the Judge within the self-play loop alongside the Challenger and the Solver, rather than keeping it fixed throughout training.
Self-evolution proceeds iteratively, with each iteration comprising three stages (Figure~\ref{fig:overview}). First, the Challenger learns to generate progressively more challenging tasks by minimizing the reward that the Judge assigns to the Solver's responses (Section~\ref{sec:challenger_solver_training}). Second, in response to these increasingly difficult tasks generated by the Challenger, the Solver is trained to produce higher-quality responses by maximizing the Judge's reward for its responses to them (Section~\ref{sec:challenger_solver_training}).
Third, the Judge is updated using the BT loss on preference pairs constructed within the self-play loop (Section~\ref{sec:judge_training}).

\subsection{Adversarial Evolution of the Challenger and Solver}
\label{sec:challenger_solver_training}

\textbf{Adversarial Challenger--Solver game.}
Let $C_{\theta_\text{c}}$ denote the Challenger, $S_{\theta_\text{s}}$ the Solver, and $J_{\phi}$ the Judge.
The Challenger samples a batch of $N$ tasks,
$\mathcal{X}=\{x_i\}_{i=1}^{N}$, where $x_i \sim C_{\theta_\text{c}}$.
For each task $x_i \in \mathcal{X}$, the Solver samples $M$ responses,
$\mathcal{Y}_i=\{y_{i,j}\}_{j=1}^{M}$, where
$y_{i,j} \sim S_{\theta_\text{s}}(\cdot \mid x_i)$.
The Judge assigns each task--response pair a scalar score
$r^{S}_{i,j}=\sigma\left(J_{\phi}(x_i,y_{i,j})\right)$,
where $\sigma(\cdot)$ maps the raw Judge output to $[0,1]$.
The Challenger and Solver interact through an asymmetric adversarial game:
\begin{equation}
\label{eq:adversarial_objectives}
\min_{\theta_\text{c}}
\mathcal{L}_{C}(\theta_\text{c};\theta_\text{s},\phi),
\qquad
\max_{\theta_\text{s}}
\mathcal{R}_{S}(\theta_\text{s};\theta_\text{c},\phi).
\end{equation}
Here, the Solver objective is determined directly by the Judge scores,
\begin{equation}
\label{eq:solver_reward_objective}
\mathcal{R}_{S}(\theta_\text{s};\theta_\text{c},\phi)
=
\mathbb{E}_{x\sim C_{\theta_\text{c}}}
\mathbb{E}_{y\sim S_{\theta_\text{s}}(\cdot\mid x)}
\left[
\sigma\left(J_{\phi}(x,y)\right)
\right],
\end{equation}
whereas the Challenger objective additionally incorporates auxiliary constraints that discourage repetitive or malformed tasks.
Specifically, we define the Challenger loss as the negative expected composite reward,
\begin{equation}
\label{eq:challenger_loss_objective}
\mathcal{L}_{C}(\theta_\text{c};\theta_\text{s},\phi)
=
-\mathbb{E}_{x_i\sim C_{\theta_\text{c}}}
\left[r^{C}_i\right],
\end{equation}
where $r^{C}_i$ is defined below.
Consequently, the interaction is adversarial but not strictly zero-sum: the Challenger seeks tasks on which the Solver performs poorly while maintaining task diversity and validity, whereas the Solver learns to obtain high Judge scores on the challenging tasks generated by the Challenger.

\textbf{Challenger reward.}
For each generated task $x_i$, the mean Judge score over the $M$ Solver responses is
\begin{equation}
\label{eq:mean_judge_reward}
\bar{r}_i
=
\frac{1}{M}\sum_{j=1}^{M}r^{S}_{i,j}.
\end{equation}
This estimates how well the current Solver handles $x_i$. We therefore define the task difficulty reward as $1-\bar{r}_i$, assigning higher rewards to tasks that the Solver cannot yet answer well.

Optimizing difficulty alone, however, may lead the Challenger to generate near-duplicate tasks or malformed outputs. Following R-Zero~\citep{huang2025rzero}, we augment the difficulty reward with a repetition penalty and a format check. To measure repetition, we compute pairwise distances $d_{pq}=1-\mathrm{BLEU}(x_p,x_q)$ and group tasks satisfying $d_{pq}<\tau$ into clusters $\{\mathcal{C}_1,\ldots,\mathcal{C}_L\}$. Each task is penalized according to the relative size of its cluster:
\begin{equation}
\label{eq:repetition_penalty}
r^{\mathrm{rep}}_i
=
\lambda\frac{|\mathcal{C}_k|}{N},
\qquad
x_i\in\mathcal{C}_k,
\end{equation}
where $\lambda$ controls the penalty strength.
For the format check, each rollout must contain a well-formed task enclosed within \texttt{<question>} tags.
The resulting composite Challenger reward is
\begin{equation}
\label{eq:challenger_reward}
    r^{C}_i =
    \begin{cases}
        \max\bigl(0,\; 1 - \bar{r}_i - r^{\mathrm{rep}}_i\bigr), & \text{if } x_i \text{ passes the format check},\\[2pt]
        -1 - r^{\mathrm{rep}}_i, & \text{otherwise}.
    \end{cases}
\end{equation}

\textbf{Challenger policy update.}
Because all $N$ tasks are sampled from the same task-generation instruction, they constitute a single GRPO group. The Challenger parameters $\theta_\text{c}$ are optimized via GRPO to maximize the composite reward in Eq.~(\ref{eq:challenger_reward}), which is equivalent to minimizing the loss in Eq.~(\ref{eq:challenger_loss_objective}):
\begin{equation}
\label{eq:challenger_obj}
\mathcal{J}_{C}(\theta_{\mathrm{c}})
=
\frac{1}{N}\sum_{i=1}^{N}
\frac{1}{|x_i|}\sum_{t=1}^{|x_i|}
\Bigl[
\min\Bigl(
\rho^{C}_{i,t}\hat{A}^{C}_i,
\operatorname{clip}
\left(
\rho^{C}_{i,t},
1-\epsilon,
1+\epsilon
\right)
\hat{A}^{C}_i
\Bigr)
-
\beta
\mathbb{D}_{\mathrm{KL}}
\left[
C_{\theta_{\mathrm{c}}}
\,\Vert\,
C_{\mathrm{ref}}
\right]
\Bigr],
\end{equation}
\begin{equation*}
\text{where}\;\; \hat{A}^{C}_i
=
\frac{r^{C}_i-\operatorname{mean}(\{r^{C}_i\}_{i=1}^N)}
{\operatorname{std}(\{r^{C}_i\}_{i=1}^N)+\varepsilon}
\;\;
\text{and} \;\;
\rho^{C}_{i,t}
=
\frac{
C_{\theta_{\mathrm{c}}}
(x_{i,t}\mid x_{i,<t})
}{
C_{\theta_{\mathrm{c}}^{\mathrm{old}}}
(x_{i,t}\mid x_{i,<t})
}.
\end{equation*}

\textbf{Task selection for Solver evolution.}
After updating the Challenger, we freeze it and sample a larger pool of candidate tasks. We retain the tasks that provide the most informative training signal for the Solver. For each candidate task $x_i$, the Solver generates $M$ responses, and the Judge assigns them scores
$\{r^{S}_{i,j}\}_{j=1}^{M}$. We measure the response-level score dispersion as
\begin{equation}
\label{eq:task_dispersion}
s_i
=
\operatorname{std}
\left(
\{r^{S}_{i,j}\}_{j=1}^{M}
\right)
\end{equation}
and select the top-$K$ tasks with the largest $s_i$.
These tasks lie near the current Solver's capability frontier, where its responses vary substantially in quality.
This criterion is grounded in recent theoretical analysis. \citet{bae2026online} proved that the expected policy improvement from training on a task is lower-bounded by the variance of its rewards, so tasks with high score dispersion are precisely those with the greatest room for learning. Our criterion can also be viewed as a continuous generalization of the informative band of R-Zero~\citep{huang2025rzero}. R-Zero relies on a binary verifier and therefore selects tasks by intermediate Solver accuracy, whereas our Judge produces continuous scores, so score dispersion serves as the analogous filter for identifying informative tasks.

\textbf{Solver policy update.}
Holding the Challenger and Judge fixed, we train the Solver on the $K$ selected tasks using GRPO.
For each task $x_i$, the Solver samples a group of $G$ responses, and each response receives the Judge-defined reward.
The Solver parameters $\theta_{\mathrm{s}}$ are then updated using the following GRPO objective:\footnote{
This objective is structurally identical to the Challenger GRPO objective in Eq.~(\ref{eq:challenger_obj}), with the distinction that it operates over Solver responses and normalizes advantages within each task-specific group of $G$ responses.
}
\begin{equation}
\label{eq:solver_obj}
\resizebox{\dimexpr\linewidth-2.5em\relax}{!}{%
$\displaystyle
\mathcal{J}_{S}(\theta_{\mathrm{s}})
=
\frac{1}{KG}
\sum_{i=1}^{K}\sum_{j=1}^{G}
\frac{1}{|y_{i,j}|}
\sum_{t=1}^{|y_{i,j}|}
\Bigl[
\min\Bigl(
\rho^{S}_{i,j,t}\hat{A}^{S}_{i,j},
\operatorname{clip}
\left(
\rho^{S}_{i,j,t},
1-\epsilon,
1+\epsilon
\right)
\hat{A}^{S}_{i,j}
\Bigr)
-
\beta
\mathbb{D}_{\mathrm{KL}}
\left[
S_{\theta_{\mathrm{s}}}
\,\Vert\,
S_{\mathrm{ref}}
\right]
\Bigr].
$%
}
\end{equation}

\vspace{-0.1cm}
Through these alternating updates, the Challenger continually expands the task frontier, while the Solver adapts to produce increasingly high-quality responses on the newly discovered tasks.

\subsection{Judge Adaptation}
\label{sec:judge_training}


To overcome the performance ceiling imposed by a fixed Judge and enable sustained self-improvement, we allow the Judge to co-evolve with the Challenger and Solver.
Although \citet{yuan2024selfrewarding} showed that self-rewarding methods can work with Judge-generated preference pairs, in which the highest-reward response is labeled chosen and the lowest-reward response is labeled rejected, this strategy risks reinforcing the Judge’s own biases.
We therefore impose two requirements on Judge co-evolution: (i) preference pairs must be constructed entirely within the closed loop, without external supervision, and (ii) their labels must not depend on signals produced by the Judge itself. To satisfy these requirements, we exploit two complementary sources of supervision that remain available even when the Judge is miscalibrated: the asymmetry between the roles of the Challenger and Solver, and the quality improvement obtained by decomposing difficult tasks into easier subtasks.

\textbf{Role-asymmetry pairs.}
For each held-out task $x$, the chosen response is sampled from the Solver, whereas the rejected response is produced by asking the Challenger to solve its own task under the same answer-generation prompt:
\begin{equation}
\label{eq:role_pair}
\left(
y^{+}_{\mathrm{role}},
y^{-}_{\mathrm{role}}
\right)
=
\left(
y^{S},
y^{C}
\right),
\qquad
y^{S}\sim S_{\theta_{\mathrm{s}}}(\cdot\mid x),
\quad
y^{C}\sim C_{\theta_{\mathrm{c}}}(\cdot\mid x).
\end{equation}

\vspace{-0.1cm}
The preference label follows directly from how the two policies are trained. The Solver is optimized to answer the generated tasks well, whereas the Challenger is optimized to make tasks difficult and receives no learning signal for answering them. Consequently, the Challenger's responses are systematically weaker: $y^{S}\succ y^{C}$.
Importantly, this ordering is induced by the policies' roles rather than by the current Judge's scores. Role-asymmetry pairs can therefore re-inject discriminative supervision in regions where the Judge is uncertain or miscalibrated. Collecting these preference pairs over the held-out tasks yields the role-asymmetry dataset $\mathcal{D}_{\mathrm{role}}
=
\left\{
\left(
x,
y^{S},
y^{C}
\right)
\;\middle|\;
x\in\mathcal{X}_{\mathrm{held\text{-}out}}
\right\}.$

\textbf{Subtask-amplification pairs.}
Although role-asymmetry pairs provide a clear preference-learning signal, relying on them alone may cause the Judge to saturate at the current Solver’s capability level, leaving it unable to recognize responses that surpass those produced by the current Solver. This, in turn, can cap the overall self-evolution process at the Solver’s existing capability.
To construct a response above that frontier, we adopt the principle of iterated amplification~\citep{christiano2018supervising}, under which a difficult task is decomposed into easier subtasks that a weak agent can solve more reliably.
This principle has been effective in both unverifiable domains~\citep{wu2021recursively} and verifiable domains~\citep{zhouleast}.

Concretely, the Challenger decomposes a held-out task $x$ into subtasks $\{q_k\}_{k=1}^{n_{x}}$, the Solver answers each subtask in the context of the original task, and the Challenger composes the resulting partial solutions:
\begin{equation}
\label{eq:subtask_amplification}
\begin{gathered}
\{q_k\}_{k=1}^{n_{x}}
=
\operatorname{Decompose}_{\mathrm{C}}(x),
\\[0.2em]
y_k^{\mathrm{sub}}
\sim
S_{\theta_{\mathrm{s}}}(\cdot\mid x,q_k),
\qquad
k=1,\ldots,n_{x},
\\[0.2em]
y^{\mathrm{amp}}
=
\operatorname{Compose}_{\mathrm{C}}
\left(
x,\{(q_k,y_k^{\mathrm{sub}})\}_{k=1}^{n_{x}}
\right).
\end{gathered}
\end{equation}

We compare the resulting amplified response with a one-shot response sampled from the same Solver:
\begin{equation}
\label{eq:amplification_pair}
\left(
y^{+}_{\mathrm{amp}},
y^{-}_{\mathrm{amp}}
\right)
=
\left(
y^{\mathrm{amp}},
y^{S}
\right),
\qquad
y^{S}
\sim S_{\theta_{\mathrm{s}}}(\cdot\mid x).
\end{equation}

Because the Solver is more reliable on the easier subtasks than on the original task as a whole, the response composed from their solutions tends to be more complete and accurate than a direct one-shot response: $y^{\mathrm{amp}}\succ y^{S}$.
These pairs therefore expose the Judge to response quality above the Solver's current one-shot frontier, allowing its evaluation capability to evolve toward the region that the Solver enters next as it improves.
Collecting these ordered response pairs over the held-out tasks yields the subtask-amplification preference dataset $\mathcal{D}_{\mathrm{amp}}
=
\left\{
\left(
x,
y^{\mathrm{amp}},
y^{S}
\right)
\;\middle|\;
x\in\mathcal{X}_{\mathrm{held\text{-}out}}
\right\}.$
Case studies of generated tasks and their Challenger-produced decompositions are provided in Section~\ref{app:decomposition-examples}.

\textbf{Bradley--Terry update.}
Let $\mathcal{D}=\mathcal{D}_{\mathrm{role}}\cup\mathcal{D}_{\mathrm{amp}}$ denote the union of the two preference-pair sets.
Starting from the Judge parameters obtained in the previous iteration, we update the Judge by minimizing the BT loss~\citep{bradley1952rank}:
\begin{equation} \label{eq:bt_loss} \mathcal{L}_{J}(\phi) = -\,\mathbb{E}_{(x,\, y^{+},\, y^{-}) \sim \mathcal{D}} \Bigl[\log \sigma\bigl(J_{\phi}(x, y^{+}) - J_{\phi}(x, y^{-})\bigr)\Bigr]. \end{equation}
Both types of preference pairs are constructed from the latest Challenger and Solver outputs. Judge training therefore focuses on the current frontier of self-evolution, where differences in response quality are the most difficult to evaluate reliably. This frontier continuously advances as the Challenger generates harder tasks and the Solver produces stronger responses. By minimizing Eq.~(\ref{eq:bt_loss}), the Judge learns to correct its misrankings on these challenging examples, enabling it to acquire evaluation capability tailored to the latest policy it supervises.


\begin{remark}[Complementary preference signals]
The two pair types are reliable at different stages of training, and their union therefore provides sustained supervision for the Judge throughout self-evolution (Section~\ref{sec:analysis_judge_data}).
\end{remark}

\section{Experiments}
\label{sec:exp}

\subsection{Experimental Setup}

\textbf{Models and baselines.}
We conduct experiments on Qwen3-4B-Base and Qwen3-8B-Base~\citep{yang2025qwen3} to assess performance across model scales.
Our baselines are the base model itself and two representative zero-data self-play frameworks, R-Zero~\citep{huang2025rzero} and G-Zero~\citep{huang2026gzero}.
We use \texttt{Skywork-Reward-V2-Llama-3.1-8B}~\citep{liu2025skyworkrewardv2} as the Judge model.
To assess generality across model families and Judge scales, we additionally evaluate \ours with Llama-3.2-3B-Instruct~\citep{grattafiori2024llama} as both Challenger and Solver while retaining the same Judge, as well as with a smaller \texttt{Skywork-Reward-V2-Llama-3.2-3B} Judge paired with Qwen3-4B-Base policies (Section~\ref{app:add_exps}). 

\textbf{Benchmarks.}
We evaluate all methods on 11 verifiable and 3 unverifiable benchmarks.
The verifiable domain set consists of 7 math reasoning benchmarks, 3 general-domain reasoning benchmarks, and IFEval~\citep{zhou2023instruction} for instruction following.
The unverifiable domain benchmarks are AlpacaEval 2.0~\citep{dubois2024length}, Arena-Hard-v2.0~\citep{li2024crowdsourced}, and EQ-Bench Creative Writing v3~\citep{creative-writing-bench-v3}.
For all methods, we stop training once the average score on either the verifiable or unverifiable domain begins to drop, and select the final checkpoint before this decline as the best checkpoint.
All evaluation protocols are listed in Section~\ref{app:benchmarks}.

\textbf{Implementation details.}
All experiments run on the \texttt{verl} framework~\citep{sheng2024hybridflow}.
In each self-evolution iteration, we train the Challenger for 5 steps, the Solver for 15 steps, and the Judge for 8 steps.
For the Challenger and the Solver, we mostly follow the hyperparameter settings used in prior work~\citep{huang2025rzero}.
Full implementation details are provided in Section~\ref{app:hyperparam}.

\subsection{Results}

\definecolor{bestblue}{RGB}{220,235,244}
\newcommand{\best}[1]{\cellcolor{bestblue}\textbf{#1}}
\newcommand{\bestcaption}[1]{%
  \begingroup
  \setlength{\fboxsep}{1.5pt}%
  \colorbox{bestblue}{#1}%
  \endgroup
}

\begin{table*}[t]
\centering

\caption{
Results across verifiable domains.
The \emph{Overall} score is the mean of the three domain averages.
\bestcaption{\textbf{Best results}} are highlighted.
}
\label{tab:main_results_verifiable}
\vspace{-0.1in}

{\small
\setlength{\tabcolsep}{5.0pt}
\renewcommand{\arraystretch}{1.18}

\begin{adjustbox}{max width=\textwidth}
\begin{tabular}{l*{8}{c}}
\toprule
\multirow{2}{*}{\textbf{Benchmark}}
&
\multicolumn{4}{c}{\textbf{Qwen3-4B-Base}}
&
\multicolumn{4}{c}{\textbf{Qwen3-8B-Base}}
\\
\cmidrule(lr){2-5}
\cmidrule(lr){6-9}
&
\makecell{\textbf{Base Model}\\(w/o training)}
&
\textbf{R-Zero}
&
\textbf{G-Zero}
&
\makecell{\textbf{\ours}\\(ours)}
&
\makecell{\textbf{Base Model}\\(w/o training)}
&
\textbf{R-Zero}
&
\textbf{G-Zero}
&
\makecell{\textbf{\ours}\\(ours)}
\\

\midrule
\multicolumn{9}{c}{\textit{Mathematical Reasoning}} \\
\midrule

GSM8K
& 86.96
& \best{92.34}
& 90.22
& 92.04
& 91.66
& \best{93.86}
& 93.33
& 92.95
\\

MATH500
& 75.60
& 77.80
& 74.80
& \best{79.80}
& 72.20
& 79.40
& 76.40
& \best{83.40}
\\

Minerva
& 45.22
& 52.57
& 47.06
& \best{54.04}
& 48.90
& 57.35
& 48.53
& \best{61.76}
\\

OlympiadBench
& 41.39
& 44.36
& 41.10
& \best{47.18}
& 40.95
& 44.96
& 44.21
& \best{53.12}
\\

AMC23
& 45.39
& 52.50
& 47.81
& \best{53.36}
& 44.92
& 56.56
& 49.77
& \best{60.62}
\\

AIME24
& 8.96
& 11.04
& 11.15
& \best{16.15}
& 10.52
& 13.96
& 12.71
& \best{19.58}
\\

AIME25
& 6.67
& 7.92
& 7.50
& \best{15.83}
& 8.96
& 12.29
& 10.83
& \best{15.94}
\\
\midrule
\textbf{Average}
& 44.31
& 48.36
& 45.66
& \best{51.20}
& 45.44
& 51.20
& 47.97
& \best{55.34}
\\

\midrule
\multicolumn{9}{c}{\textit{General Reasoning}} \\
\midrule

MMLU-Pro
& 51.70
& 55.55
& 54.14
& \best{58.60}
& 58.97
& 60.92
& 59.44
& \best{63.80}
\\

SuperGPQA
& 26.53
& 28.63
& 27.56
& \best{29.35}
& 30.45
& 31.87
& 31.24
& \best{33.22}
\\

BBH
& 50.88
& 64.35
& 58.90
& \best{70.85}
& 66.21
& 71.31
& 66.15
& \best{78.38}
\\

\midrule
\textbf{Average}
& 43.04
& 49.51
& 46.87
& \best{52.93}
& 51.88
& 54.70
& 52.28
& \best{58.47}
\\

\midrule
\multicolumn{9}{c}{\textit{Instruction Following}} \\
\midrule

Prompt Strict
& 40.11
& 42.33
& 40.85
& \best{50.65}
& 46.40
& 50.46
& \best{51.57}
& 49.72
\\

Instruction Strict
& 51.08
& 54.20
& 52.64
& \best{60.91}
& 58.15
& 61.63
& \best{63.19}
& 62.71
\\

Prompt Loose
& 43.99
& 48.43
& 47.32
& \best{57.86}
& 51.76
& 57.12
& 54.90
& \best{61.92}
\\

Instruction Loose
& 54.32
& 59.23
& 58.03
& \best{66.55}
& 62.47
& 67.03
& 66.19
& \best{73.02}
\\

\midrule
\textbf{Average}
& 47.38
& 51.05
& 49.71
& \best{58.99}
& 54.70
& 59.06
& 58.96
& \best{61.84}
\\
\midrule
\textbf{Overall Avg.}
& 44.91
& 49.64
& 47.41
& \best{54.38}
& 50.67
& 54.99
& 53.07
& \best{58.55}
\\
\bottomrule
\end{tabular}
\end{adjustbox}
}
\end{table*}
\begin{table*}[t]
\centering
\caption{
Results across unverifiable domains.
The \emph{Overall} score is the mean of the three benchmark scores, where the two Arena-Hard subsets are first averaged.
\bestcaption{\textbf{Best results}} are highlighted. \textit{H.P.} and \textit{C.W.} denote Hard Prompt and Creative Writing, respectively. 
}
\label{tab:main_results_unverifiable}

\vspace{-0.1in}

{\small
\setlength{\tabcolsep}{5.0pt}
\renewcommand{\arraystretch}{1.18}

\begin{adjustbox}{max width=\textwidth}
\begin{tabular}{l*{8}{c}}
\toprule
\multirow{2}{*}{\textbf{Benchmark}}
&
\multicolumn{4}{c}{\textbf{Qwen3-4B-Base}}
&
\multicolumn{4}{c}{\textbf{Qwen3-8B-Base}}
\\
\cmidrule(lr){2-5}
\cmidrule(lr){6-9}
&
\makecell{\textbf{Base Model}\\(w/o training)}
&
\textbf{R-Zero}
&
\textbf{G-Zero}
&
\makecell{\textbf{\ours}\\(ours)}
&
\makecell{\textbf{Base Model}\\(w/o training)}
&
\textbf{R-Zero}
&
\textbf{G-Zero}
&
\makecell{\textbf{\ours}\\(ours)}
\\

\midrule

AlpacaEval
& 6.22
& 11.38
& 9.20
& \best{28.56}
& 12.93
& 18.37
& 18.39
& \best{33.53}
\\

Arena-Hard (H.P.)
& 2.50
& 2.50
& 3.00
& \best{4.80}
& 4.00
& 5.70
& 4.40
& \best{6.90}
\\

Arena-Hard (C.W.)
& 0.90
& 1.50
& 1.40
& \best{2.20}
& 1.70
& 2.20
& 2.20
& \best{3.90}
\\

EQ-Bench C.W.
& 20.83
& 24.59
& 21.26
& \best{30.36}
& 23.92
& 24.30
& 24.25
& \best{31.31}
\\

\midrule
\textbf{Overall Avg.}
& 9.58
& 12.66
& 10.89
& \best{20.81}
& 13.23
& 15.54
& 15.31
& \best{23.41}
\\

\bottomrule

\end{tabular}
\end{adjustbox}
}
\end{table*}

We evaluate \ours in the verifiable (Table~\ref{tab:main_results_verifiable}) and the unverifiable domains (Table~\ref{tab:main_results_unverifiable}).
\ours attains the best score on every benchmark group at both scales.

\textbf{Verifiable domain.}
\ours improves the average performance in the verifiable domain by \textbf{9.47} and \textbf{7.88} points over the corresponding base models (Qwen3-4B-Base and Qwen3-8B-Base, respectively), while outperforming R-Zero by \textbf{4.74} and \textbf{3.56} points. Notably, \ours surpasses R-Zero, even though R-Zero is specifically designed for self-evolution in verifiable domains.

\textbf{Unverifiable domain.}
Baselines achieve much smaller gains in the unverifiable domain compared to the verifiable one, and this is where the gap to \ours is widest.
R-Zero relies on a majority-vote reward that does not extend to unverifiable open-ended tasks, so its average improves by only 3.08 and 2.31 points, roughly half of what it gains on the verifiable side.
G-Zero achieves even smaller gains of 1.31 and 2.08 points, which leaves it behind R-Zero and barely above the base model, since G-Zero does not employ the Judge at all. 
\ours improves the average performance in the unverifiable domain by \textbf{11.23} and \textbf{10.18} points, respectively, with the largest gains observed on AlpacaEval 2.0 (6.22 $\rightarrow$ 28.56 and 12.93 $\rightarrow$ 33.53), a broad general instruction-following benchmark that covers diverse open-ended tasks across areas such as writing, business communication, personal advice, planning, and recommendations, while also including tasks in mathematics and factual knowledge.


\section{Analysis}
\label{sec:analysis} 


\subsection{Reliability of Self-Generated Preference Labels}
\label{sec:analysis_judge_data}

\begin{wrapfigure}{t}{0.46\textwidth}
\centering
\vspace{-0.2in}
\includegraphics[width=\linewidth]{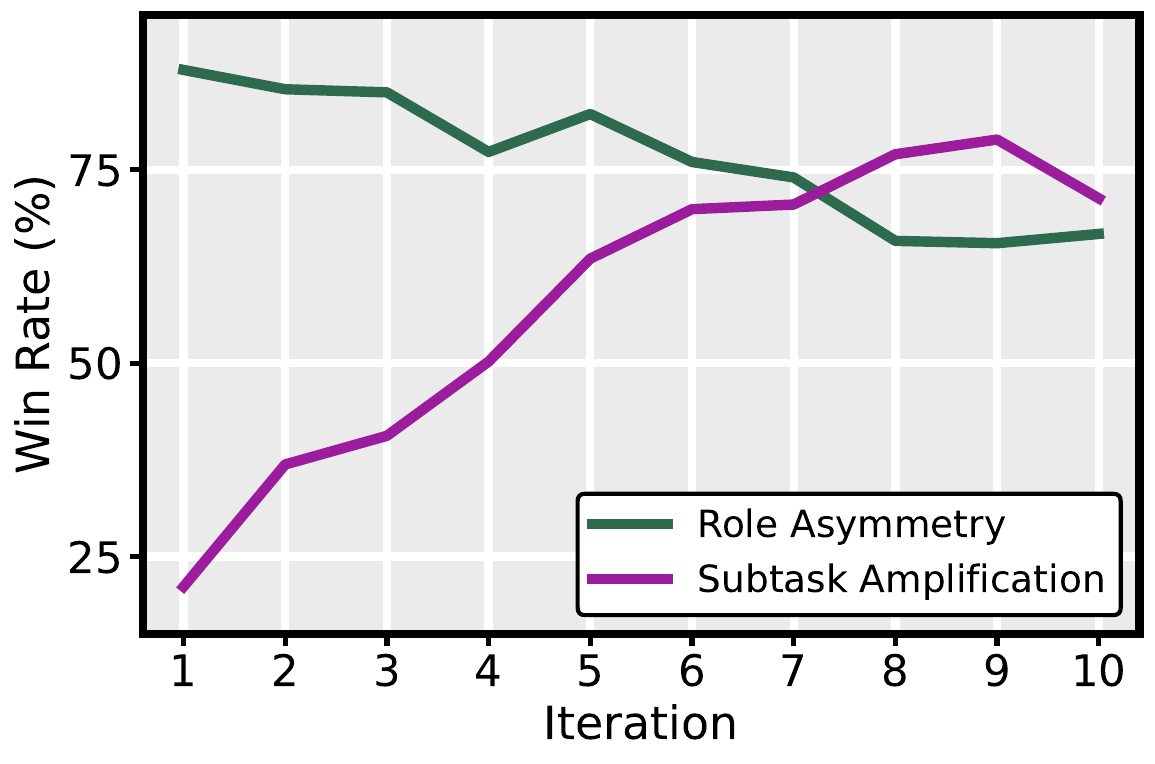}
\vspace{-0.28in}
\caption{
Win rate of the chosen response for the two Judge-training pair types at each iteration.
Each comparison is evaluated in both presentation orders, with ties excluded.
}
\label{fig:winrate_iter}
\vspace{-0.1in}
\end{wrapfigure}

\ours assumes that the preference labels generated within the loop are reliable, so we evaluate their correctness directly.
At each iteration of the Qwen3-4B-Base experiments\footnote{Unless otherwise noted, we report all analyses using Qwen3-4B-Base.}, we present both responses from every Judge-training pair to an external LLM judge, \texttt{Claude Opus 4.8}~\citep{opus4p8}, and ask it to identify the better response.
We then measure how often the side we label as chosen wins.
To mitigate positional bias, we evaluate both presentation orders and drop ties (Figure~\ref{fig:winrate_iter}). The judging instruction is provided in Section~\ref{app:winrate_instruction}.

\textbf{Role-asymmetry pairs are reliable from the start.}
The Solver's response wins more than 60\% of the comparisons at every iteration, so the labels in $\mathcal{D}_{\mathrm{role}}$ agree with an independent evaluator throughout training.
The win rate decreases from $87.9\%$ to approximately $66\%$. We attribute this decline to the increasingly difficult adversarial curriculum rather than to unreliable preference pairs. Because the Challenger is rewarded for generating tasks that the Solver struggles to answer, the held-out tasks gradually shift toward the limits of the Solver's capabilities. On these difficult tasks, both the Challenger and Solver struggle to produce strong responses, so the quality gap between them becomes smaller.

\textbf{Subtask-amplification pairs become reliable once the Solver matures.}
The divide-and-conquer response wins fewer than half of the comparisons in the first three iterations ($21.1\%$ at iteration 1), since decomposition pays off only once the Solver can reliably solve the subtasks.
From iteration 4, the win rate exceeds 50\%, and it later reaches roughly 70 to 80\%, which confirms that $\mathcal{D}_{\mathrm{amp}}$ supplies supervision above the Solver's one-shot frontier.
The two curves cross in the middle of training, so the Judge is never left without a usable signal.
Role-asymmetry pairs carry the signal early, and amplification pairs take over once the Solver matures.

While our primary goal is to make the Judge adaptive to the current Solver rather than to improve its standalone evaluation capability, we also find that Judge co-evolution improves performance on RM-Bench~\citep{liu2025rm}, an independent reward-model benchmark unrelated to the preference pairs constructed within the self-evolution loop (Section~\ref{app:judge_rmbench}).


\subsection{Ablation Study on Preference Data for Judge Adaptation}
\label{sec:ablation}
We conduct an ablation study to measure how much each type of Judge training data contributes to \ours (Table~\ref{tab:ablation_study}).
Both components, role-asymmetry and subtask-amplification pairs, contribute to the gains over the fixed-Judge baseline, with subtask amplification providing the larger improvement. Removing subtask-amplification pairs leads to a larger performance drop ($-1.64$) than removing role-asymmetry pairs ($-0.97$). This is consistent with the design of the two pair types. Role-asymmetry pairs teach the Judge to distinguish responses below the Solver’s current capability, whereas subtask-amplification pairs expose it to responses beyond the Solver’s one-shot frontier, which is critical for continued improvement. Nevertheless, combining the two sources performs best because they are most reliable at different stages of training (Section~\ref{sec:analysis_judge_data}).


\subsection{Sustained Improvement Over Iterations}
\label{sec:lifelong}

\begin{figure}[h!]
    \centering
    \includegraphics[width=1\linewidth]{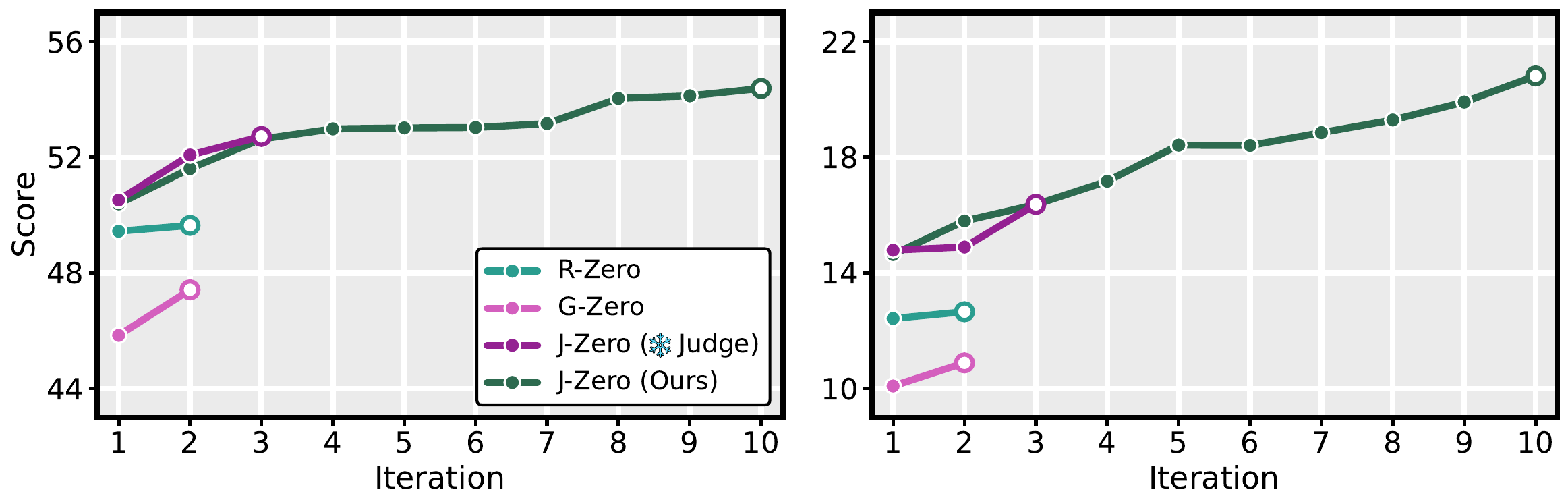} 
    \caption{Average score per iteration on the verifiable (\textbf{left}) and unverifiable (\textbf{right}) benchmarks. Each method is plotted up to its best checkpoint.}
    \label{fig:analysis_lifelong}
\end{figure}

\begin{wraptable}{t}{0.5\textwidth}
\centering
\small
\vspace{-0.15in}
\caption{Ablation results. We disable one component at a time.}
\label{tab:ablation_study}
\vspace{-0.1in}

\resizebox{\linewidth}{!}{%
\begin{tabular}{@{}lccc@{}}
    \toprule
    \textbf{Method}
    & \textbf{Verifiable}
    & \textbf{Unverifiable}
    & \textbf{Overall} \\
    \midrule
    \ours
    & \textbf{54.38}
    & \textbf{20.81}
    & \textbf{37.59} \\
    \midrule
    \enspace $\vdash$
    \raisebox{-0.3ex}{%
        \includegraphics[width=0.04\linewidth]{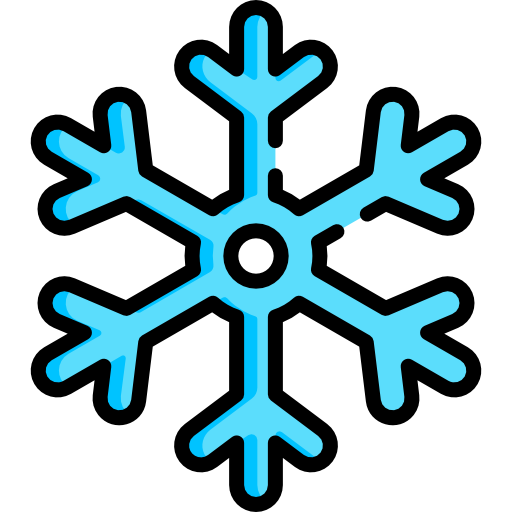}%
    }
    Judge
    & 52.72
    & 16.37
    & 34.54 \\
    \midrule
    \enspace $\vdash$ w/o $\mathcal{D}_{\mathrm{amp}}$
    & 53.20
    & 18.71
    & 35.95 \\
    \enspace $\vdash$ w/o $\mathcal{D}_{\mathrm{role}}$
    & 53.68
    & 19.55
    & 36.62 \\
    \bottomrule
\end{tabular}%
}
\vspace{-0.1in}

\end{wraptable}
\textbf{\ours does not plateau within our training budget.}
Existing LLM self-play methods plateau after only a few iterations: R-Zero and
G-Zero peak at iteration 2 and decline thereafter.
\ours instead improves monotonically through iteration 10 in both
domains, gaining 9.47 points and 11.23 points over the
base model on the verifiable and unverifiable benchmarks, respectively
(Figure~\ref{fig:analysis_lifelong}).

\textbf{Judge co-evolution is what sustains improvement.}
The frozen-Judge variant follows \ours closely for the first three iterations and
then plateaus, ending 1.66 and 4.44 points below the full run on the verifiable and unverifiable domains, respectively (Table~\ref{tab:ablation_study} and Figure~\ref{fig:analysis_lifelong}). The divergence point is informative: it is where the Solver reaches the fixed Judge's own evaluation ceiling, after which the Judge's rewards no longer separate good responses from bad. A co-evolving Judge keeps its evaluation standard above the Solver's current level, so the reward signal stays discriminative as the Solver improves.

\section{Concluding Remarks}
\label{sec:conclusion}

\textit{Conclusion.} We have introduced \ours, a zero-data self-evolving framework in which the Challenger, Solver, and Judge co-evolve to support self-improvement in both verifiable and unverifiable domains. Role asymmetry provides reliable preference signals early in training, while subtask amplification supplies supervision beyond the Solver’s one-shot capability. Together, these signals allow the Judge to overcome a fixed evaluation ceiling and keep pace with the evolving policies without external data or human feedback. Empirically, \ours outperforms prior zero-data methods at two different model scales (Table~\ref{tab:main_results_verifiable},~\ref{tab:main_results_unverifiable}) and continues to improve through ten iterations, whereas existing approaches saturate within two (Figure~\ref{fig:analysis_lifelong}). These findings establish the Judge as a critical trainable component: a self-evolving model can improve only as far as its evaluator can see.


\textit{Limitations.} Compute constraints limit us to Challenger and Solver policies of up to 8B parameters with an 8B Judge, and to base models only; larger scales and post-trained reasoning models that emit long chains of thought remain untested. Our Judge is also a classifier-based discriminative reward model, initialized from an off-the-shelf checkpoint and trained with the BT loss, whereas the Challenger and Solver share a single generative initialization. A generative Judge (\eg LLM-as-a-judge) would let one base model instantiate all three roles, and its critiques could serve as richer in-loop supervision; how to make such a generative Judge co-adapt with the Challenger and Solver within the self-play loop is an interesting direction for future work.

\bibliographystyle{plainnat}
\bibliography{paper}

\clearpage
\appendix{\section*{Appendix}
\label{sec:appendix}

\section{Full Experimental Details}
\label{app:exp_details}

\subsection{Implementation Details}
\label{app:hyperparam}

\begin{table}[!h]
\centering
\small
\caption{Full hyperparameter settings.}
\label{tab:full_hyperparameter_setting}
\begin{tabular}{l|c|c|c}
    \toprule
    \textbf{Hyperparameters}
    & \textbf{Challenger}
    & \textbf{Solver}
    & \textbf{Judge} \\
    \midrule

    Steps per iteration
        & 5
        & 15
        & 8 \\
    Training batch size
        & 16
        & 128
        & 64 \\
    Mini-batch size
        & 16
        & 32
        & -- \\

    \midrule

    \multirow[c]{2}{*}{Max length}
        & Prompt: 1024
        & Prompt: 4096
        & \multirow[c]{2}{*}{8192} \\
        & Response: 4096
        & Response: 4096
        & \\

    \midrule

    Learning rate
        & $1 \times 10^{-6}$
        & $1 \times 10^{-6}$
        & $5 \times 10^{-7}$ \\
    LR scheduler
        & constant
        & constant
        & constant \\
    Weight decay
        & 0.0
        & 0.0
        & 0.0 \\
    KL penalty coefficient
        & 0.01
        & 0.01
        & -- \\

    \midrule

    Number of rollouts
        & 4
        & 5
        & -- \\
    Rollout temperature
        & 1.0
        & 1.0
        & -- \\
    Rollout top-$p$
        & 0.99
        & 0.99
        & -- \\
    Clip ratio
        & $(0.20, 0.28)$
        & $(0.20, 0.28)$
        & -- \\

    \bottomrule
\end{tabular}
\end{table}

\paragraph{Training setup.}
We implement all experiments on top of the \texttt{verl} framework~\citep{sheng2024hybridflow} and conduct all training on four NVIDIA B200 GPUs and four NVIDIA H200 GPUs.
The detailed hyperparameter settings are provided in Table~\ref{tab:full_hyperparameter_setting}.
Unless otherwise noted, \ours follows the hyperparameters of R-Zero~\citep{huang2025rzero}; in particular, we use a Challenger training batch size of 16, as in the official R-Zero implementation\footnote{\url{https://github.com/Chengsong-Huang/R-Zero}}, and turn off weight decay.
The prompts used for the Challenger are in Section~\ref{app:prompts_chal}, and the prompts used for the Solver are in Section~\ref{app:prompts_sol}.
During Judge training, we use equal proportions of role-asymmetry and subtask-amplification preference pairs.

\paragraph{Baseline configurations.}
Beyond the settings common to all methods, we keep each baseline's own configuration.
For R-Zero, we use the same Challenger training batch size of 16 as \ours.
For G-Zero~\citep{huang2026gzero}, we keep the LoRA-based training setup and the Challenger training batch size of 128 from the original work.
In preliminary experiments, both replacing LoRA with full-parameter fine-tuning and reducing the Challenger training batch size to 16 lowered performance.

\subsection{Benchmarks and Evaluation}
\label{app:benchmarks}

\paragraph{Math reasoning benchmarks.}

We evaluate all methods on 7 benchmarks: GSM8K~\citep{cobbe2021gsm8k}, MATH500~\citep{hendrycksmath2021}, Minerva~\citep{lewkowycz2022minerva}, OlympiadBench~\citep{he2024olympiadbench}, AMC23, AIME24, and AIME25. Following R-Zero~\citep{huang2025rzero}, we set \texttt{"Please reason step by step, and put your final answer within \textbackslash boxed\{\}."} as a system prompt, obtain responses with up to 4096 tokens, and report the avg@32 for AMC and AIME, whereas greedy decoding accuracy is reported for the remaining benchmarks. We use \texttt{math\_verify} as a primary verifier and also employ Qwen3.6-27B as an additional judge to semantically verify the answer.

\paragraph{General domain benchmarks.}

We evaluate all methods on 3 benchmarks: MMLU-Pro~\citep{wang2024mmlu}, SuperGPQA~\citep{du2026supergpqa}, and Big-Bench Hard~\citep[BBH;][]{suzgun2023challenging}. We obtain responses with up to 8192 tokens and report accuracy with greedy decoding. We mostly follow R-Zero, but slightly strengthen the evaluation code to prevent false positives, where an incorrect answer is randomly marked as correct.

\paragraph{Instruction-following benchmarks.}

To evaluate instruction-following capabilities, we use IFEval~\citep{zhou2023instruction}. Following the official evaluation source code\footnote{\url{https://github.com/google-research/google-research/tree/master/instruction_following_eval}}, we report all four metrics: prompt-level strict accuracy, instruction-level strict accuracy, prompt-level loose accuracy, and instruction-level loose accuracy.

\paragraph{Unverifiable domain benchmarks.}

We evaluate on 3 benchmarks: AlpacaEval 2.0~\citep{dubois2024length}, Arena-Hard-v2.0~\citep{li2024crowdsourced}, and EQ-Bench Creative Writing v3~\citep{creative-writing-bench-v3}.
For AlpacaEval 2.0, we report length-controlled win rate against GPT-4-Turbo.
For Arena-Hard-v2.0, we report win rates on both the Hard Prompt subset (with style control) against o3-mini and the Creative Writing subset against gemini-2.0-flash.
For EQ-Bench Creative Writing v3, we report rubric score instead of Elo rating to avoid model-pool dependence and align the metric scale with other benchmarks.
We use Qwen3.6-27B as a judge for AlpacaEval 2.0 and Arena-Hard-v2.0 (Hard Prompt). For Arena-Hard-v2.0 (Creative Writing) and EQ-Bench Creative Writing v3, we adopt gemma-4-31B-it~\citep{gemma4} as a judge since it achieves a higher EQ-Bench Judgemark v4\footnote{\url{https://eqbench.com/judgemark-v4.html}} score than Qwen3.6-27B, indicating stronger discriminative performance in creative writing evaluation.
Except for judge models, all benchmarks are evaluated following the official protocols and configurations.

\section{Additional Experimental Results}
\label{app:add_exps}

\subsection{Performance on a Different Policy Model Family}
\label{app:jzero_llama}

\begin{wraptable}{t}{0.5\textwidth}
\centering
\small
\vspace{-0.15in}
\caption{Results on Llama-3.2-3B-Instruct.}
\label{tab:app_results_llama}
\vspace{-0.1in}

\begin{tabular}{@{}lccc@{}}
    \toprule
    \textbf{Method}
    & \textbf{Verifiable}
    & \textbf{Unverifiable}
    & \textbf{Overall} \\
    \midrule
    Base Model & 44.13 & 16.97 & 30.55 \\
    \midrule
    R-Zero & 44.91 & 17.05 & 30.98 \\
    \midrule
    G-Zero & 43.94 & 16.20 & 30.07 \\
    \midrule
    \ours
    & \textbf{46.02}
    & \textbf{17.32}
    & \textbf{31.67} \\
    \bottomrule
\end{tabular}

\vspace{-0.1in}
\end{wraptable}

To further evaluate our method, we conduct additional experiments using Llama-3.2-3B-Instruct~\citep{grattafiori2024llama}. We follow the same training configuration and evaluation protocol described in Appendix~\ref{app:exp_details}, and compare R-Zero, G-Zero, and \ours against the initial checkpoint. We use \texttt{Skywork-Reward-V2-Llama-3.1-8B} as the Judge model, consistent with the main experiments.

\ours achieves the highest average performance on both verifiable and unverifiable tasks, with scores of 46.02 and 17.32, respectively, outperforming the strongest baseline, R-Zero, by 1.11 and 0.27 points (Table~\ref{tab:app_results_llama}). These results suggest that the benefits of \ours generalize beyond the Qwen family to other model architectures.

\subsection{Performance on a Smaller Judge}
\label{app:jzero_3brm}

\begin{wraptable}{t}{0.5\textwidth}
\centering
\small
\vspace{-0.15in}
\caption{Comparison with a smaller Judge. \ours \raisebox{-0.3ex}{%
    \includegraphics[width=0.04\linewidth]{icon/snowflake.png}%
} denotes \ours without Judge adaptation. The 8B scores are from Table~\ref{tab:ablation_study}.}
\label{tab:app_3b8b}
\vspace{-0.1in}

\resizebox{\linewidth}{!}{%
\begin{tabular}{@{}llccc@{}}
    \toprule
    \textbf{Judge}
    & \textbf{Method}
    & \textbf{Verifiable}
    & \textbf{Unverifiable}
    & \textbf{Overall} \\
    \midrule

    8B
    & \ours \raisebox{-0.3ex}{%
        \includegraphics[width=0.04\linewidth]{icon/snowflake.png}%
    }
    & 52.72
    & 16.37
    & 34.54 \\

    &
    \ours
    & \textbf{54.38}
    & \textbf{20.81}
    & \textbf{37.59} \\

    \midrule

    3B
    & \ours \raisebox{-0.3ex}{%
        \includegraphics[width=0.04\linewidth]{icon/snowflake.png}%
    }
    & 52.06
    & 19.58
    & 35.82 \\

    &
    \ours
    & \textbf{52.62}
    & \textbf{22.33}
    & \textbf{37.48} \\

    \bottomrule
\end{tabular}%
}

\vspace{-0.2in}
\end{wraptable}

We additionally evaluate our method with the \texttt{Skywork-Reward-V2-Llama-3.2-3B}~\citep{liu2025skyworkrewardv2} Judge while keeping the Challenger and Solver initialized from Qwen3-4B-Base. At each Judge size, we compare \ours and its frozen-Judge variant to assess the effectiveness of Judge co-evolution.

\ours achieves an overall score of 37.48 with the 3B Judge, only 0.11 points below the 8B Judge (Table~\ref{tab:app_3b8b}). This small difference accompanies a domain-level trade-off: using the 3B Judge achieves a lower verifiable average (52.62 vs. 54.38) but a higher unverifiable average (22.33 vs. 20.81). Notably, Judge co-adaptation improves both the verifiable and unverifiable averages over the corresponding frozen variant at both scales, yielding overall gains of 1.66 points with the 3B Judge and 3.05 points with the 8B Judge. These results demonstrate that a smaller Judge can also achieve competitive downstream performance, while the benefits of Judge co-adaptation remain evident at both evaluated Judge scales.

\section{Judge Improvements}
\label{app:judge_rmbench}

The purpose of Judge co-adaptation in \ours is to keep the supervision appropriate for the \emph{latest} Solver, not to turn the Judge into a stronger reward model on static benchmarks.
Nevertheless, it is natural to ask whether the co-adapted Judge also becomes better in absolute terms.
To address this, we evaluate the Judge from every iteration on RM-Bench~\citep{liu2025rm} (Figure~\ref{fig:judge_rmbench}).

\textbf{The Judge improves in every domain.}
Accuracy improves in all four RM-Bench domains and the average rises monotonically from 92.61 to 93.95 (+1.34).
The largest improvement is in Chat (+3.70), the domain closest to the open-ended tasks that the Challenger writes, followed by Math (+1.24) and Code (+0.34).
Safety is already saturated at iteration 0 (98.69) and remains nearly unchanged (+0.10).
Thus, even while adapting to the loop, the Judge does not lose its general reward-modeling capability but rather improves slightly.

\begin{figure}[t]
    \centering
    \includegraphics[width=1\linewidth]{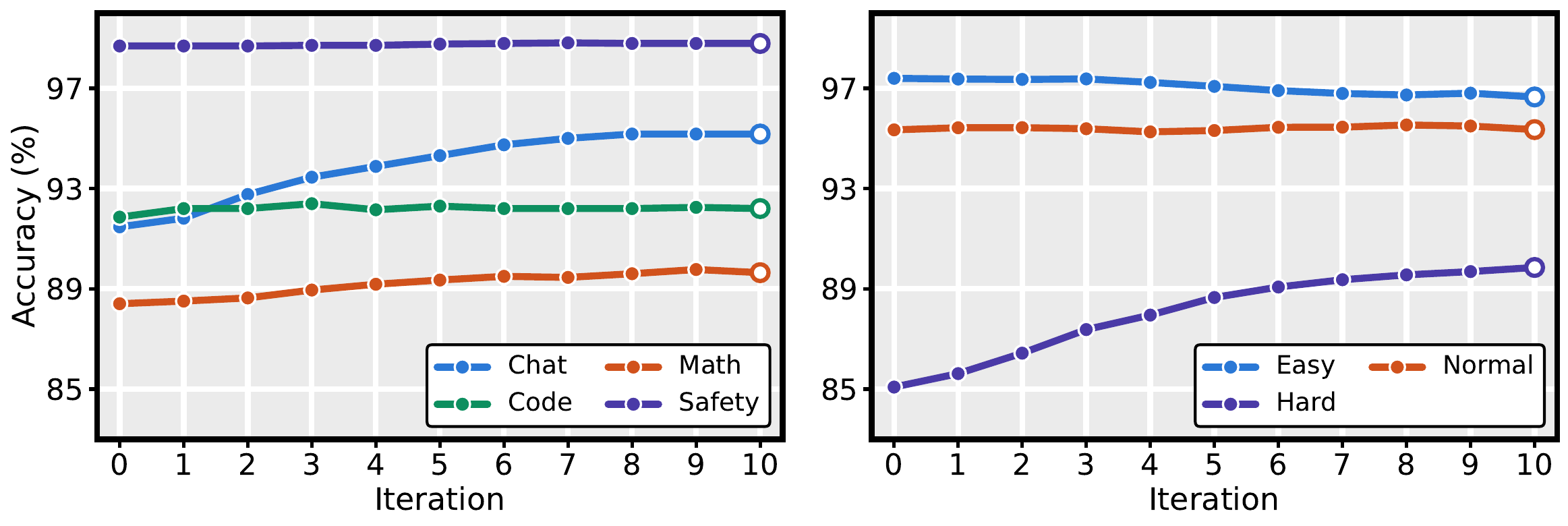}
    \caption{Judge performance on RM-Bench per iteration. Accuracy on each domain subset (\textbf{left}) and accuracy at each difficulty level (\textbf{right}).}
    \label{fig:judge_rmbench}
\end{figure}

\textbf{The gains concentrate on hard preference pairs.}
Accuracy on Hard pairs increases by 4.77 points, from 85.08 to 89.85, while Normal is unchanged (+0.01) and Easy decreases by 0.74 points, from 97.40 to 96.66.
The three levels differ in whether response style agrees with response quality: an Easy pair presents the better response in the more elaborate style, whereas a Hard pair presents it more plainly than the worse one.
We do not read the small decline on Easy as a meaningful loss.
The reason is that separating a clearly good response from a clearly bad one carries less information as training proceeds, since a strong Solver rarely produces an obviously bad response.
Residual preference for surface polish, by contrast, is directly exploitable: a fluent but substantively flawed response can receive a high score and be reinforced on that bias, and the Hard split is what measures this failure.
Hard comparisons are the ones the Judge must resolve correctly for an already strong policy to keep improving, and they are where that accuracy improves most.
Over self-evolution the Judge becomes a stronger reward model, with the gain concentrated at the difficulty level where the Solver places the greatest demand on it.



\newtcolorbox{promptbox}[1][]{
  enhanced,
  breakable,
  colback=gray!5,
  colframe=black!70,
  boxrule=0.6pt,
  arc=1.5mm,
  left=2mm, right=2mm, top=1.5mm, bottom=1.5mm,
  title=\textbf{Prompt},
  fonttitle=\small,
  coltitle=white,
  colbacktitle=black!70,
  before skip=8pt, after skip=8pt,
  pad at break*=1mm,
  #1
}

\section{Prompts}
\label{app:prompts}

\subsection{Challenger Prompts in \ours}
\label{app:prompts_chal}
This section contains every prompt used by the Challenger. We do not set a system prompt for problem generation; the Challenger only receives the user prompt. The prompts follow the order of the pipeline. First is the user prompt for problem generation, modified from the original prompt used in G-Zero~\citep{huang2026gzero}. Next are the system and user prompts for task decomposition, which we use to build the subtask-amplification pairs that are used to train the Judge. Last are the system and user prompts that compose the Solver's per-subtask responses into a single answer.

\begin{promptbox}[title=Challenger User Prompt for Generating Questions (Modified from G-Zero)]
Produce one challenging request that a real user might ask a capable assistant.\\
\\
The request should come from a general-domain distribution. Sample across task types, not from a single area. Examples of task types you can draw from:\\
  - writing (email, story, essay, pitch, review, poem)\\
  - explanation (make a concept clear to a specific audience)\\
  - advice or planning (career, travel, project, learning)\\
  - analysis (argument, text, dataset description, product)\\
  - coding (small function, debugging, design question)\\
  - role-play, dialogue, or creative tasks\\
  - open-ended questions about ethics, science, everyday life\\
  - reasoning, math, or logic problems (fine to include - roughly 1 in 6 requests, no more)\\
\\
Weight the non-math categories above heavily. A little math is good for diversity, but it should not dominate - favor tasks where the response quality depends on tone, structure, audience-fit, clarity, or creativity, not just arithmetic correctness.\\
\\
Requirements:\\
- The request must be self-contained and non-trivial to answer well.\\
- Wrap the request in <question> and </question> tags.\\
- Output nothing else before, between, or after the tagged blocks.\\
\\
Example 1 (writing):\\
\textless question\textgreater Write a resignation email to my manager that keeps the door open for future collaboration. I've been at the company for 4 years and I'm leaving to join a competitor. Tone should be professional and warm without being effusive.\textless/question\textgreater\\
\\
Example 2 (explanation):\\
\textless question\textgreater Explain what a Kalman filter does to a software engineer who is comfortable with linear algebra but has never touched signal processing. Avoid control-theory jargon where possible.\textless/question\textgreater\\
\\
Now produce one new request of your own:
\end{promptbox}

\begin{promptbox}[title=Challenger System Prompt to Decompose a Task for Subtask-amplification Pair]
You are an expert at breaking a hard task into simpler subtasks. Given a task, output the FEW essential sub-questions/subtasks whose answers, once combined, fully solve the task.\\
Rules:\\
- Output the 3-5 most essential subtasks only — never more than 5.\\
- Each subtask must be SIMPLER than the whole task and genuinely needed.\\
- Order them so earlier subtasks are useful for later ones.\\
- Make them specific to THIS task; avoid vague meta-steps like `understand the problem' or `write the answer'.\\
Output a numbered list of the subtasks only, nothing else.
\end{promptbox}

\begin{promptbox}[title=Challenger User Prompt to Decompose a Task for Subtask-amplification Pair]
Task: \{\{\textit{Given Task}\}\}\\
Break this into the 3-5 essential, simpler subtasks that together solve it, ordered so earlier ones help later ones.
\end{promptbox}

\begin{promptbox}[title=Challenger System Prompt to Compose the Sub-answers for Subtask-amplification Pair]
You are an expert problem solver. You are given a task, and a set of subtasks that have each already been solved. Combine the subtask solutions into ONE coherent, complete final answer to the original task: integrate them, resolve any contradictions, fix obvious subtask errors, and do not be unnecessarily verbose.\\
Output ONLY the final answer to the original task — do NOT restate the subtasks, their solutions, or your procedure. If the task has a closed-form answer, put it within \textbackslash boxed\{\}.
\end{promptbox}

\begin{promptbox}[title=Challenger User Prompt to Compose the Sub-answers for Subtask-amplification Pair]
Original task: \{\{\textit{Given Task}\}\}\\
Solved subtasks: \{\{\textit{Sub-answers}\}\}\\
Combine these into the single best final answer to the original task. Output ONLY the final answer.
\end{promptbox}

\subsection{Solver Prompts in \ours}
\label{app:prompts_sol}
This section lists the prompts used by the Solver. When the Solver answers a problem, its user prompt is the question generated by the Challenger, so the system prompt is the only one we set at that stage; it appears first below. The two prompts after it, a system prompt and a user prompt, are the ones we use to collect the Solver's response to each subtask when building the subtask-amplification pairs.

\begin{promptbox}[title=Solver System Prompt]
Please reason step by step, and put your final answer within \textbackslash boxed\{\}.
\end{promptbox}

\begin{promptbox}[title=Solver System Prompt to Solve Each Subtask for Subtask-amplification Pair]
You are solving ONE subtask that is part of a larger task. Answer the subtask fully and correctly, using the larger task only as context. Be concise and self-contained - output only the answer to this subtask.
\end{promptbox}

\begin{promptbox}[title=Solver User Prompt to Solve Each Subtask for Subtask-amplification Pair]
Larger task (context): \{\{\textit{Given Task}\}\}\\
Subtask to solve now: \{\{\textit{Given Sub-task}\}\}\\
Answer this subtask.
\end{promptbox}

\subsection{LLM-as-a-judge Prompt for Judge Training Data Analysis}
\label{app:winrate_instruction}

The prompt used for Judge training data analysis in Section~\ref{sec:analysis_judge_data} is shown below.

\begin{promptbox}[title=LLM-as-a-judge Prompt for Judge Training Data Analysis]
You are an impartial expert judge. Each item has a QUESTION and two candidate answers, **A** and **B**, produced by different models for the SAME question. Decide which answer is better. **Forced choice: exactly one winner, "A" or "B" - never a tie.**\\
\\
\#\# Step 1 - classify the question's domain\\
One of: \textasciigrave math | code | explain | advice | writing | other\textasciigrave\\
\\
\#\# Step 2 - judge by that domain's primary criteria\\
- **math**: correctness of the final answer dominates - work it out yourself and check; then validity/clarity of the reasoning.\\
- **code**: does it correctly satisfy the request (would it run / meet the spec)? completeness and edge cases; then readability.\\
- **explain**: factual accuracy first; then completeness, clarity, structure.\\
- **advice**: usefulness and specificity to the actual situation, actionability, sound judgment; generic boilerplate loses to targeted, concrete guidance.\\
- **writing**: satisfies ALL stated constraints (form, length, topic, tone), coherence, craft, originality.\\
- **other**: instruction-following, accuracy, overall helpfulness.\\
\\
\#\# Universal rules\\
- **Degeneration loses heavily.** Treat as degenerate: repetition loops; prompt/question echoed back instead of answered; empty or near-empty output; scaffolding or meta-text leaking into the answer (e.g. "system:", "User:", "Assistant:", "Subtask 1:", role/turn markers, composition or grading instructions); self-cut-off mid-sentence.\\
- \textasciigrave ...[TRUNCATED]\textasciigrave at the very end is OUR display cutoff, not the model's - judge only what is shown and do NOT penalize it.\\
- Judge substance, not length. Longer is not better.\\
- Do not try to guess which system produced A or B; judge only the text in front of you. A and B were already position-randomized.\\
\\
\#\# Degeneration flags\\
Independently of who wins, set \texttt{deg\_a} / \texttt{deg\_b} to true if that specific answer is degenerate by the rule above.\\
\\
\#\# Output - STRICT\\
Emit one JSON object per item, one per line (JSONL), no prose, no markdown fences:\\
\\
\textasciigrave \textasciigrave \textasciigrave\\
\{"id":"\textless item id\textgreater", \\
"domain":"math | code | explain | advice | writing | other",\\
"winner":"A", \\
"deg\_a":false, \\
"deg\_b":false,\\
"reason":"one short sentence"\}\\
\textasciigrave \textasciigrave \textasciigrave
\end{promptbox}

\section{Subtask Amplification Examples}
\label{app:decomposition-examples}

This appendix presents representative examples of user prompts and the
corresponding subtasks produced by the decomposition step.


\begin{promptbox}
Provide a project plan for a team of software developers tasked with
designing and implementing a collaborative office space monitoring
system. The system should include features such as real-time occupancy
tracking, ambient noise levels, air quality monitoring, and energy
usage analysis. The plan should be detailed with key milestones,
estimated timelines, and roles and responsibilities. Also, outline the
necessary resources, such as hardware, software libraries, and
external API integrations, and include a short description of how user
data will be handled according to GDPR standards.
\end{promptbox}

\noindent\textbf{Subtasks:}
\begin{enumerate}[itemsep=0pt, topsep=2pt]
  \item Define project scope and requirements
  \item Assign roles and responsibilities
  \item Identify and procure necessary resources
  \item Develop and test system features
  \item Ensure GDPR compliance in data handling
\end{enumerate}


\begin{promptbox}
Develop a comprehensive travel itinerary for a week-long trip to
Paris, France. The itinerary should include transportation options
from New York to Paris, lodging recommendations in major tourist and
cultural hotspots, and a detailed list of activities tailored for a
family with two children aged 8 and 10, focusing on a mix of
historical sites, culinary experiences, and outdoor adventures.
Provide the itinerary in a markdown format, with each day's schedule
clearly outlined and suggestions for dining and accommodations.
\end{promptbox}

\noindent\textbf{Subtasks:}
\begin{enumerate}[itemsep=0pt, topsep=2pt]
  \item Research and select the most cost-effective and efficient
        transportation options from New York to Paris.
  \item Identify and recommend suitable hotels for each day of the
        trip, prioritizing families with children.
  \item Plan daily activities that cater to a family with children,
        balancing historical, culinary, and outdoor interests.
  \item Suggest local restaurants and cafes that offer family-friendly
        dining options.
  \item Organize the itinerary in a markdown format, ensuring each
        day's schedule is clear and easy to follow.
\end{enumerate}


\begin{promptbox}
Design a detailed marketing plan for a new coffee shop brand that has
been around for only three months. Consider both the unique aspects of
the coffee shop offering (such as locally sourced beans and specialty
brews) and high-level marketing strategies, including social media
strategy, target audience segmentation, partnerships with local
businesses, pricing strategy, and even the opening of a second
location after six months.
\end{promptbox}

\noindent\textbf{Subtasks:}
\begin{enumerate}[itemsep=0pt, topsep=2pt]
  \item Define the unique selling proposition (USP) of the coffee
        shop, focusing on its local sourcing and specialty brews.
  \item Segment the target audience based on demographics,
        preferences, and behaviors relevant to coffee shop patrons.
  \item Develop a pricing strategy that reflects the USP and target
        audience's willingness to pay.
  \item Outline a social media strategy that promotes the USP and
        engages with the target audience.
  \item Identify potential partnerships with local businesses that can
        mutually benefit from each other's customer base.
\end{enumerate}


\begin{promptbox}
Imagine you are a renowned historical fiction author. Write a
character sketch for a mysterious British aristocrat who lives in the
late 18th century, avoids public appearances, and is rumored to be a
Freemason. The protagonist should be between 30--40 years old, with a
pale complexion, sharp features, and a knack for espionage. Include
details about their personality, motivations, and a back story that
ties them to revolutionary French politics. The tone should be elegant
and intricately detailed.
\end{promptbox}

\noindent\textbf{Subtasks:}
\begin{enumerate}[itemsep=0pt, topsep=2pt]
  \item Define the character's physical appearance.
  \item Establish the character's age and social status.
  \item Describe the character's personality traits and motivations.
  \item Create a compelling backstory related to revolutionary French
        politics.
  \item Craft the tone and style of the character sketch.
\end{enumerate}



\begin{promptbox}
As a software developer, help me design a small function in Python
that calculates the factorial of a given number. The function should
include error handling for negative inputs and should be optimized for
performance. Additionally, provide a brief explanation of how the
factorial calculation works and why the error handling is important in
this context.
\end{promptbox}

\noindent\textbf{Subtasks:}
\begin{enumerate}[itemsep=0pt, topsep=2pt]
  \item Define the function with a clear name, such as
        \texttt{factorial}, and specify the input parameter (an
        integer).
  \item Implement the factorial calculation using a loop or recursion,
        optimized for performance.
  \item Include error handling for negative inputs by checking the
        parameter and raising a custom exception.
  \item Provide a brief explanation of how the factorial calculation
        works (the math concept + applications).
  \item Explain the importance of error handling here (consequences of
        not handling negatives).
\end{enumerate}

}

\end{document}